\documentclass[runningheads]{packages/llncs}
\usepackage{graphicx}

\usepackage{xcolor}
\usepackage{hyperref}

\usepackage{amsmath}
\usepackage{amssymb}
\usepackage{mathtools}
\usepackage[nameinlink,capitalize]{cleveref}

\newcommand{\bm}[1]{\boldsymbol{#1}}

\newcommand{\Conv}{\mathop{\scalebox{1.7}{\raisebox{-0.2ex}{$\ast$}}}}%
\newcommand{\set}[1]{\left\lbrace #1 \right\rbrace}
\newcommand{\abs}[1]{\left\lvert #1 \right\lvert}
\newcommand{\norm}[1]{\left\lVert #1 \right\lVert}

\newcommand{\cs}{\mathcal C}

\newcommand{\depen}[1]{^{(#1)}}
\newcommand{\bhh}{\bm {\hat H}}

\newcommand{\rh}{\bm \rho\depen{\bm H}}

\newcommand{\rhh}{\bm \rho\depen{\bhh}}

\newcommand{\R}{{\mathbb R}}
\newcommand{\X}{{\mathcal X}}
\newcommand{\Y}{{\mathcal Y}}
\newcommand{\C}{{\mathcal C}}
\newcommand{\E}{{\mathbb E}}
\newcommand{\N}{{\mathbb N}}
\newcommand{\Z}{{\mathbb Z}}
\newcommand{\I}{{\mathcal I}}

\newcommand{\calH}{{\mathcal H}}

\newcommand{\Hcnn}{{\mathcal H}_{\text{CNN}}}
\newcommand{\Hcnnl}{{\mathcal H}_{\text{L-CNN}}}

\begin{document}
\title{Forward and Inverse Approximation Theory for Linear Temporal Convolutional Networks}
\author{Haotian Jiang\inst{1} \and
Qianxiao Li \inst{1, 2}}

\institute{Department of Mathematics, National University of Singapore \and
Institute for Functional Intelligent Materials, National University of Singapore
\email{qianxiao@nus.edu.sg}}
\maketitle              % typeset the header of the contribution
\renewcommand*{\proofname}{Sketch of proof}
%  \pagenumbering{gobble}
\begin{abstract}
    % In our previous study \cite{jiang2021.ApproximationTheoryConvolutionala}, 
    % we conducted a theoretical analysis of the effectiveness of convolutional architectures when applied to the modeling of temporal sequences.
    % This work aims to improve upon previous research by presenting novel results. 
    % Specifically, introduce a revised complexity measure for the targets, resulting in tighter approximation rates. 
    % We also provide inverse approximation theorems based on this complexity measure. 
    % Taken together, these results provide a comprehensive characterization of the hypothesis spaces of convolutional networks as applied to the modeling of temporal relationships.
    We present a theoretical analysis of the approximation properties of convolutional architectures when applied to the modeling of temporal sequences.
    Specifically, we prove an approximation rate estimate (Jackson-type result)
    and an inverse approximation theorem (Bernstein-type result),
    which together provide a comprehensive characterization of
    the types of sequential relationships that
    can be efficiently captured by a temporal
    convolutional architecture.
    The rate estimate improves upon a previous result
    via the introduction of a refined complexity measure,
    whereas the inverse approximation theorem is new.
    \keywords{Approximation  \and Temporal convolutional neural networks \and Sequence Modeling.}
\end{abstract}

\section{Introduction}
Although convolutional neural networks (CNNs) are commonly used for image inputs, 
they are shown to be effective in tackling a range of temporal sequence modeling problems compared to the traditional recurrent neural networks \cite{yin2017.ComparativeStudyCNN,bai2018.EmpiricalEvaluationGeneric}.
For example, WaveNet \cite{oord2016.WaveNetGenerativeModel} is a convolution-based model for generating audio from text. 
It utilizes a multilayer dilated convolutional structure, resulting in a filter with a large receptive field.
The approximation properties of the dilated convolutional architectures applied to sequence modeling have been studied in \cite{jiang2021.ApproximationTheoryConvolutionala}. 
A complexity measure is defined to characterize the types of targets that can be efficiently approximated.

This work aims to enhance prior research and contribute new findings. 
We formulate the approximation of sequence modeling in a manner that parallels classic function approximation, where we consider three distinct types of approximation results, including universal approximation, approximation rates, and inverse approximation. 
Our main contributions are summarized as follows:  
\textbf{1.}  We refine the complexity measure to make it naturally adapted to the approximation of convolutional architectures. 
The resulting approximation rate in the forward approximation theorem is tighter compared to results in \cite{jiang2021.ApproximationTheoryConvolutionala}.
\textbf{2.} We prove a Bernstein-type inverse approximation result.
It states that a target can be effectively approximated only if its complexity measure is small, 
which presents a converse of the forward approximation theorem.

 The readers may refer to \cite{jiang2023.BriefSurveyApproximation} for a detailed discussion of related research on the approximation theory for sequence modeling.

% \clearpage
% \thispagestyle{empty}
\section{Mathematical Formulation of sequence modeling problems}
This section provides a theoretical formulation of sequence modeling problems and defines the architecture of the dilated convolutional model. Specifically, a temporal sequence is defined as a function $\bm x: \I \to \R^d$ that maps an index set $\I$ to real vectors. The value of the sequence at time index $t$ is denoted by $x(t) \in \R^d$. Additionally, a finitely-supported discrete sequence can also be considered as a matrix, where each column denotes a time step and the total number of columns is referred to as the length of the filter.

% In this section, we theoretically formulate the sequence modeling problems and define the dilated convolutional model architecture.
% A temporal sequence $\bm x: \I \to \R^d$ is regarded as a function from the index set $\I$ to real vectors.
% The value at time index $t$ is denoted by $x(t) \in \R^d$.
% We may also consider a finitely-supported discrete sequence as a vector.
% We use $\bm x_{[t_1,t_2]}$ to denote a truncation of $\bm x$ in the interval $[t_1,t_2]$. 
% For a discrete sequence $\bm \rho: \mathbb N \to \mathbb R^d$, define the radius of $\bm \rho$ by 
% $\rad{\bm \rho} = \sup\set{ s:\rho(s) \neq 0}$.
% The interval $[0, r(\bm \rho)]$ contains non-zero parts of $\bm \rho$, which we refer as its support.
% We use $\abs{\cdot}$ to denote the Euclidean norm of a vector.

\subsubsection{Sequence to sequence modeling}
In this study, we consider the supervised learning problem of sequence to sequence (seq2seq) modeling.
That is, given an input $\bm x$, we want to predict a corresponding output sequence $\bm y$ at each time step $t$.
We use $\mathbb{Z}$ to represent the set of integers and $\mathbb{N}$ to represent the set of non-negative integers.
For an output at time $t \in \Z$, the relationship between the output $y(t)\in\R$ and $\bm x$ can be considered as a functional $H_t$ such that
$
    y(t) = H_t(\bm x).
$
Thus, the mapping between two temporal sequences $\bm x$ and $\bm y$ can be regarded as a sequence of functionals $\bm H = \set{H_t: t\in\Z}$.
As an example, let's consider the shift of sequences. Given $\bm x$, the corresponding output $\bm y$ is a shift of $\bm x$ in time such that $y(t) = x(t-k)$. Then this sequence to sequence mapping can be written as $y(t) = H_t(\bm x) = \sum_{s = -\infty}^\infty \rho(s)x(t-s)$, where $\rho(s) = 1$ when $s=k$ and $\rho(s)=0$ otherwise.

We now specify the input and output spaces. 
We use $\bm c(\Z, \mathbb R^d)$ to denote the set of temporal sequences from the integer index set $\Z$ to $d$ dimensional real vectors, and use $\bm c_0$ to denote temporal sequences with compact support.
The input space is defined as 
$
	\mathcal X = \Big\{ \bm x \in  \bm c(\mathbb Z, \mathbb R^d):
	\sum_{s \in \mathbb Z} \abs{x(s)}^2 < \infty \Big\}.
$
This is indeed the $\ell^2$ sequence space.
It forms a Banach space under the norm $\norm{\bm x}_{\mathcal X}^2 := { \sum_{s\in \mathbb Z} \abs{x(s)}^2}$.
For the output space, we consider scalar temporal sequences, 
$
    \Y = \bm c(\Z,\R).
$
% To deal with vector-valued outputs, we can consider each dimension separately.

\subsubsection{Dilated convolution architecture}
% We now define the dilated convolutional architectures.
% \begin{definition}
	Let $\bm f: \mathbb Z \to \mathbb R^d$, $\bm g: \mathbb Z \to \mathbb R^d$ be two discrete sequences.  The discrete causal dilated convolution with dilation rate $r$ is defined as 
	$
		(\bm f \Conv{}_{r} \, \bm g)(t) = \sum_{s\geq 0}f(s)^\top g(t-r s).
	$
	When $r=1$, this is the usual convolution.
    % , and we denote it by $\Conv$.
% \end{definition}
A general dilated temporal CNN model with
$K$ layers and $M$ channels at each layer is
\begin{equation}\label{eq:CNNdynamics}
		\begin{aligned}
		 \bm h_{1,i}
        &=
        \sum_{j=1}^d\bm w_{0ji} \Conv{}_{1}\bm x_j,  \quad 
		 \bm h_{k+1,i} = \sigma \left(\sum_{j=1}^{M} \bm  { w}_{kji}
         \Conv{}_{d_k} \bm h_{k,j} \right), \\
        % &=
        % \sigma \left(\sum_{j=1}^{M_k} \bm  { w}_{kji}
        % \Conv{}_{d_k} \bm h_{k,j} \right),\\
		\bm {\hat y} &=  \sum_{i=1}^M\bm h_{K,i},
		\end{aligned}
\end{equation}
where $i=1, \dots, M$, $k=1, \dots, K-1$.
% Here, $\bm x_{j}$ denotes the scalar sequence corresponding to the $j^\text{th}$ dimension of $\bm x$.
Furthermore, ${\bm w}_{kji}:\N\to\R$ is the convolutional filter at layer $k$,
mapping from channel $j$ at layer $k$ to channel $i$ at layer $k+1$, 
which has compact support $[0,l-1]$.
We assume all the filters ${\bm w}_{kji}$ have the same length $l\geq 2$.
We set the dilation rate to be $d_k = l^K$, 
which is the standard practice for dilated convolution architecture \cite{oord2016.WaveNetGenerativeModel,yu2016.MultiScaleContextAggregation}.
The element-wise activation function is denoted as $\sigma$.
The dilated CNN hypothesis space is defined as follows
\begin{equation}
    \Hcnn =
    \bigcup_{K, M}
    \Hcnn^{(M,K)}
    =
    \bigcup_{K, M}
    \Big\{
        \bm{x} \mapsto \bm{\hat{y}}
        \text{ in \cref{eq:CNNdynamics}}
    \Big\}.
\end{equation}
\section{Approximation results}

In this section, we present the approximation results.
We consider a linear setting where the activation $\sigma$ is linear.
Despite the linearity of the activation function, 
the dependence on time remains nonlinear. 
As shown in \cite{li2022.ApproximationOptimizationTheory} and  \cite{jiang2021.ApproximationTheoryConvolutionala},
this particular setting is meaningful in that it captures the intrinsic structure of the architectures.
% in the sense that it captures the intrinsic structure of different architectures.
The CNN hypothesis space with linear activation is defined as 
\begin{equation}\label{eq:linearhcnn}
    \begin{aligned}
        \Hcnnl = \bigcup_{K, M}
        \Hcnnl^{(M,K)} = &\Big\{\bm{\hat{H}}:
        \hat{H}_t(\bm x) = \sum_{s=0}^\infty {\rho}\depen{\bhh}(s)^\top x(t-s)
        \Big\},
    \end{aligned}
\end{equation}
where $\rhh:[0,l^K-1]\to\R$ is defined by
\begin{equation}\label{eq:cnn_representation}
    \rhh_i
    =
    \sum_{i_1, \dots, i_{K}=1}^M
    \bm{w}_{K-1, i_{K-1}, i_K}
    \Conv{}_{l^{K-1}}
    % \bm{w}_{K-2, i_{K-2}, i_{K-1}}
    % \Conv{}_{l^{K-2}}
    \dots
    \Conv{}_{l^1}
    \bm{w}_{0, i, i_1}.
\end{equation}
It is important to note that due to the non-associativity of dilated convolution, 
the order of the convolutions mentioned above should be evaluated from left to right.
In this work, we consider target seq2seq mappings satisfying the following conditions, 
which are standard in functional analysis.
\begin{definition}\label{def:target space}
    Let $\bm H = \set{H_t:t\in\mathbb Z}$ be a sequence of functionals.
    \begin{enumerate}
        \item $\bm H$ is causal if it does not depend on the future inputs:
        for any $\bm x_1,  \bm x_2 \in \mathcal X$ and any $t \in \mathbb Z$ such that
        $
            x_1(s) = x_2(s) \ \text{ for all } s \leq t
        $,
        we have $H_t(\bm x_1) = H_t(\bm x_2)$.
    
        \item For all t, $H_t \in \bm H$ is a continuous linear functional if $H_t$ is continuous and for any $\bm x_1,  \bm x_2 \in \mathcal X$
        and $\lambda_1, \lambda_2 \in \mathbb R$,
            $
            H_t(\lambda_1\bm x_1 + \lambda_2\bm x_2) = \lambda_1 H_t(\bm x_1) + \lambda_2 H_t(\bm x_2).
            % \norm{H_t} := \sup_{\bm x \in \mathcal X, \norm{\bm x}_{\mathcal X}\leq 1}\abs{H_t(\bm x)} < \infty,
            $
        % where $\norm{H_t}$ denotes the induced functional norm. 
        % The norm of a sequence of functionals is defined by $\norm{\bm H}:= \sup_{t \in \mathbb Z}\norm{H_t}$.
        % \jht{may need to modify this norm}.
        \item $\bm H$ is time-homogeneous if 
        for any $t, \tau \in \mathbb Z$,
        $
            H_t(\bm x) = H_{t+\tau}(\bm x\depen{\tau})
        $
        where $x\depen{\tau}(s) := x(s-\tau)$ for all $s\in \mathbb Z$.
    \end{enumerate}
    % \qlr{Why not use $\bm H$ here, and after?}
    \end{definition}
    
    Following \cref{def:target space}, we define the target space as
    \begin{equation}
        \begin{aligned}
        \mathcal C = \{ \bm H: \bm H \text{ satisfies conditions in \cref{def:target space}}\}.
    \end{aligned}
    \end{equation}
The following lemma is a result of Riesz representation theorem for Hilbert space \cite{kreyszig1989.IntroductoryFunctionalAnalysis},
which shows that the target space admits a convolutional representation.

\begin{lemma}\label{lmm:representation}
For any $\bm H \in \cs$,
    there exists a unique $\ell^2$ sequence $\rh :\mathbb N \to \mathbb R^d$ such that 
    \begin{equation}\label{eq: integral representation}
    	H_t(\bm x) = \sum_{s=0}^{\infty} \rho\depen{\bm H}(s)^\top x(t-s), \quad t \in \mathbb Z.
    \end{equation}
\end{lemma}
For a target $\bm H$,
the corresponding representation is denoted as $\rh$.
Compared to the hypothesis space in equation \cref{eq:linearhcnn}, 
we note that approximating $\bm H$ with CNNs is equivalent to approximating $\rh$ with $\rhh$.
The hypothesis space in this study is linked to the concept of linear time-invariant systems (LTI systems). 
Specifically, a causal LTI system will produce a linear functional that adheres to the conditions outlined in Definition \ref{def:target space}. 
It is important to note, however, 
that not all linear functionals that satisfy Definition \ref{def:target space} necessarily correspond to an LTI system. 
The objective of our investigation is to examine general functionals without assuming that the data is generated by a hidden linear system.

\subsubsection{Tensor structure of dilated convolution architectures} 
To begin with, we discuss the tensor structures related to dilated convolution architectures.
Dilated convolutional architectures are designed with the goal of achieving long filters with few parameters.
A model with $K$ layers and one channel produces a filter with length $l^K$ using only $lK$ parameters.
We are interested in how the number of channels affects the expressiveness of the model, and in this regard,
we observe that the dilated convolution architectures indeed possess a tensor structure.
By treating the filters ${w_{kji}}$ in \cref{eq:cnn_representation} as length $l$ vectors,
we replace the convolution operation with tensor products, resulting in a tensor
    \begin{equation}\label{eq:tensor_representation}
        T(\rhh_i)
        =
        \sum_{i_1, \dots, i_{K}=1}^M
        \bm{w}_{K-1, i_{K-1}, i_K}
        \otimes
        \bm{w}_{K-2, i_{K-2}, i_{K-1}}
        \otimes
        \dots
        \otimes
        \bm{w}_{0, i, i_1},
\end{equation}
where $T(\rhh_i) \in\R^{l\times\cdots\times l}$.
From this point of view, 
the number of channels $M$ determines the rank of the tensor $T(\rhh_i)$.
This inspires us to reshape the target representation $\rh$ into a tensor and consider the approximation between tensors,
where this approach was first introduced in our previous work \cite{jiang2021.ApproximationTheoryConvolutionala}
We make use of higher-order singular value decomposition (HOSVD) \cite{delathauwer2000.MultilinearSingularValue} to achieve this.
According to HOSVD, a tensor $\bm A \in \R^{I_1 \times \cdots \times I_K}$ can be decomposed into
$
    \bm  A = \sum_{i_1}^{I_1} \cdots \sum_{i_K}^{I_K}s_{i_1\dots i_K} \,\, \bm u_{i_1} \otimes \cdots\otimes
    \bm u_{i_K},
$
where $s_{{i_1} \dots {i_K}}\in\R$ are scalars and $\bm u_{i_k} \in \R^{I_k} $
forms a set of orthonormal bases for $\R^{I_k}$. 
We are now ready to discuss the approximation results.

\subsection{Jackson-type and Bernstein-type results}
The universal approximation property (UAP) of $\Hcnnl$ is proved in  \cite{jiang2021.ApproximationTheoryConvolutionala},
which ensures that the hypothesis space $\Hcnnl$ is dense in the target space $\C$.
However, it does not guarantee the quality of the approximation. 
Specifically, given a target $\bm H \in \Hcnnl$, 
we are concerned with how the approximation error behaves as we increase $K$ and $M$. 
This is precisely what the approximation rate considers. 
As a demonstration, we use $m$ to denote the complexity of a candidate in $\Hcnnl$. 
This complexity is typically measured by the number of parameters needed, also called the approximation budget.

In general, a hypothesis $\calH = \bigcup_m\calH\depen{m}$ is usually built up by candidates with different approximation budgets, 
where a larger $m$ typically results in better approximation quality.
% The UAP ensures that the hypothesis space is dense in the target space.
% However, it does not guarantee the quality of the approximation.
% For instance, given a target $\bm H \in \calH$,
% we are concerned with how the approximation error behaves as we increase $K, M$.
% This is exactly what approximation rate is considering.
% As a demonstration,
% we use $m$ to denote the complexity of a candidate in $\calH$.
% It usually measure the number of parameters needed,
% which is also called approximation budget.
% A hypothesis $\calH = \bigcup_m\calH\depen{m}$ is usually built up by candidates with different approximation budget, 
% larger $m$ results in better approximation quality. 
We may write an approximation rate in the following form
\begin{align}
    \inf_{\hat H\in\calH\depen{m}}\norm{H- \hat H}\leq C_{\calH}(H,m).
\end{align}
The error bound $C_\calH(\cdot,m)$ decreases as $m\to\infty$, 
and the speed of decay is the approximation rate, which quantitatively shows how the error behaves. 
$C_{\calH}(H, \cdot)$ is a complexity measure for the target $H$. 
If $C_{\calH}(H, \cdot)$ decays rapidly, the target $H$ can be easily approximated.
As an example, let's consider the approximation of $f \in C^{\alpha}[0,1]$ using polynomials. 
The Jackson theorem \cite{achieser2013.TheoryApproximation} states that $C_{\calH}(f,m) = \frac{c_\alpha}{m^\alpha}\max_{r=1\dots\alpha}\lVert{f^{(r)}}\rVert_\infty$. 
This implies that a function can be efficiently approximated by a polynomial if it has a small Sobolev norm. We call these kinds of approximation rates Jackson-type results.

Jackson-type results are considered forward approximation results, 
as they allow us to determine the approximation rate given the complexity measure. 
However, the approximation rate is usually an upper bound on the error. 
We may ask whether the target can be approximated faster than the given rate.
This leads us to the inverse approximation problem, where we are given the approximation rate of a target and want to determine its complexity. 
In the context of polynomial approximation, 
the Bernstein theorem \cite{achieser2013.TheoryApproximation} states that only functions at least $\alpha$ times differentiable can be approximated faster than 
$\frac{A}{m^{\alpha}}$. 
Bernstein-type results are useful for determining the suitability of an architecture for a particular problem theoretically.
It is worth mentioning that the choice of complexity measure usually depends on the hypothesis space.
In order to investigate the approximation with dilated convolutional architectures,
we need to define suitable complexity measures.

\subsubsection{Complexity measure}
Based on the previous discussion, we define the complexity measure of targets suitable for the hypothesis space $\Hcnnl$.
Considering the spectrum of $T(\rh)$, we define the following complexity measure
\begin{align}
    C\depen{g}_1(\bm H) = \inf\left\{  
        c: \sum_{j=1}^d \sum_{i=s}^{l^K}|s_{i}\depen{j,K}|^2 \leq c g(s-1), s\geq 0, K\geq 1
    \right\},
\end{align}
where $|s_{i}\depen{j,K}|$ comes from the HOSVD of $T((\rh_{j })_{[0,l^K-1]})$ in decreasing order, and $g\in\bm c(\N,\R_{+})$ is a non-increasing function converges to zero at infinity.
This complexity measure takes into account the decay of the spectrum of the target.
Likewise,
define a complexity measure considering the tail sum of $\rh$
\begin{equation}
    C\depen{f}_2(\bm H) = \inf\left\{  
        c: \sum_{i=s}^{\infty}|\rho\depen{\bm H}(i)|^2  \leq c f(s), s\geq 0
    \right\}.
\end{equation}
We restrict ourselves to considering targets that satisfy the above two complexity measures.
\begin{align}
    \C\depen{g,f} = \{ \bm H\in \C: C_1\depen{g}(\bm H) + C_2\depen{f}(\bm H) < \infty \}.
\end{align}
Having established the complexity measures, we next present the approximation results.

\begin{theorem}[Jackson-type]\label{thm:jackson}
    Fix $l\geq 2$ and $g,f \in \bm c_0(\N,\R_{+})$. For any $\bm H \in \C\depen{g,f}$ and any $M, K$ we have
    \begin{align}
        \inf_{\bhh \in \Hcnnl\depen{M,K}} \norm{\bm H - \bhh}^2\leq C_1\depen{g}(\bm H)g(M) + C_2\depen{f}(\bm H)f(l^K).
    \end{align}
    \begin{proof}
        For $\bm x \in \X$  with $\norm{\bm x}\leq 1$, we have
        \begin{align}\label{eq:two parts error}
            \abs{H_t(\bm x)  - \hat H_t(\bm x)}^2 
            \leq 
            % \sum_{s=0}^\infty \Big\rvert \rh(s) - \rhh(s) \Big\rvert^2\nonumber
            % \\
              \sum_{s=0}^{l^K-1}\Big\rvert \rh(s) - \rhh(s)\Big\rvert^2
            + \sum_{s=l^K}^{\infty} \Big\rvert \rh(s) \Big\rvert ^2. 
        \end{align}
       For the second term of the error, we have  
        $
            \sum_{s=l^K}^{\infty} \rvert \rh(s) \rvert ^2 \leq C_2\depen{f}(\bm H)f(l^K)
        $ directly from the definition of $C_2$.
        While for the first term, we have 
        \begin{align}
            \sum_{s=0}^{l^K-1}\Big\rvert \rh(s) - \rhh(s)\Big\rvert^2& =\sum_{j=1}^d \norm{T(\rh_{j}) - T(\rhh_j)}_F^2 \\
                    &=\sum_{j=1}^d \sum_{i=M+1}^{l^K}|s_{i}\depen{j,K}|^2\leq C_1\depen{g}(\bm H)g(M).
        \end{align}
        We construct $\rhh$ such that it matches the bases of the HOSVD decomposition of $\rh$, 
        then the second equality results from the orthogonality of the bases in the HOSVD decomposition.

    \end{proof}
\end{theorem}
This theorem shows that a target can be efficiently approximated by a dilated CNN if it has a fast decaying spectrum (small $C_1$) and fast decaying memory (small $C_2$).
It is notable that the two terms in the rate are independent of each other, 
where $K$ controls the tail error and $M$ controls the error resulting from the spectrum. 
This is considered an improvement over the previous Jackson-type result presented in \cite{jiang2021.ApproximationTheoryConvolutionala}, 
where the first term has the form $C(\bm H)g(KM^{\frac 1 K}-K)$.
In that case, 
the first term increases with $K$, which may result in the overall error bound not decreasing as $K$ increases.
The current result solves this problem, 
ensuring that the overall error bound always decreases as the number of parameters increases.
Moreover, under the same $g$, the bound in our current work is tighter than the one in \cite{jiang2021.ApproximationTheoryConvolutionala}.
Next, we present the Bernstein-type result.

\begin{theorem}[Bernstein-type]\label{thm:Bernstein}
    Let $\mathcal D$ is a distribution such that at every time step $t$ we have $x(t)\sim \mathcal N(0,1)$.
    Fix $l\geq 2$ and $g,f \in \bm c_0(\N,\R_{+})$. Given $\bm H \in \C\depen{g,f}$ and $A, B >0$, suppose for any $M, K$ there exists $\bhh\depen{M,K} \in \Hcnnl\depen{M,K}$ such that
    \begin{align}
        \sup_t\mathop{\E}_{\bm x \sim \mathcal D} [|H_t(\bm x) - \hat H_t(\bm x)|^2]\leq Ag(M) + Bf(l^K),
    \end{align}
    then we have $\bm H \in C\depen{g,f} $ and $C_1\depen{g}(\bm H) \leq A$, $C_2\depen{f}(\bm H) \leq B$.
    \begin{proof}
        Since the above inequality holds for any $(K, M)$, we can again consider the two terms separately.
        For any $K$, consider a model $\bhh$ with a large $M$ such that $Ag(M)$ goes to zero.
        We conclude that for any $K$, there exists $\bhh$ such that 
        \begin{align}
            \sum_{s=l^K}^\infty \abs{\rh(s)}^2 = \lVert{\bm H - \bhh}\rVert^2 \leq Bf(l^K),
        \end{align}
        which implies $C_2\depen{f}(\bm H)\leq B$.
        The intuition of the proof is that in $[l^K, \infty]$, the error is completely determined by the tail error of $\rh$.
        Thus, the error rate directly implies the decay rate of the tail.
        Similarly, for any $M$, we pick a large enough $K$ such that $Bf(l^K)$ goes to zero.
        Then, there exists $\bhh$ such that
        \begin{align}
            \sum_{j=1}^d\sum_{s=M+1}^\infty |s_{i}\depen{j,K}|^2 = \lVert{\bm H - \bhh}\rVert^2 \leq Ag(M),
        \end{align}
        which implies $C_1\depen{g}(\bm H) \leq A$. 
    \end{proof}
\end{theorem}
The Bernstein-type result serves as a converse to the Jackson-type \cref{thm:jackson}.
It suggests that a target can be well approximated by the model only if it has small complexity measure, 
which in turn implies that the target exhibits good spectrum regularity and rapid decay of memory.
Combining the two results, 
we obtain a complete characterization of $\Hcnnl$:
a target in $\calH$ can be efficiently approximated by: linear temporal CNNs if and only if it has fast decaying
spectrum and memory. 

% We numerically verify the results as shown in \cref{fig1}.
% Multiple targets with different complexity measures are generated.
% For targets whose training error does not fall below the error bound, the corresponding complexity measures exceed $A$.
% Conversely, if the training error falls below the error bound, the complexity measure is less than $A$.

\section{Related works}

Tensor decomposition of convolutional filters  have also been discussed in 
\cite{chu2021.LowrankTensorDecomposition,hayashi2019.ExploringUnexploredTensor,kim2016.CompressionDeepConvolutional,phan2020.StableLowrankTensor}.
In these works, the problem considered is convolution applied to image inputs with shape  $(H,W,C)$. 
At a specific layer, the convolution kernel applies to it have shape $(I,J,C,C')$, 
where $I,J$ is the size of the convolution kernel and $C, C'$ is input channels and output channels, respectively. 
They mainly focus on decomposition of the four dimensional convolutional kernel, 
which is different from what we do in this current work and the previous work \cite{jiang2021.ApproximationTheoryConvolutionala}. 
Instead of considering decomposition of a kernel at a specific layer,
we consider the tensor structure of the entire model resulting from the multilayer dilated convolutions.
In our case, the convolution kernel for sequence modeling is a one dimensional array with length $l^K$.
We tensorize it into a $K$ dimensional tensor and consider its decompositions.
This enables us to analyze the precise relationship between the target and the convolutional model. 

% \begin{figure}
%     \centering
%     \includegraphics[width=0.7\textwidth]{images/output.png}
%     \caption{First, we fix $g=\exp{(-M)}$ and $A>0$, and proceed to plot $Ag(M)$ as the error bound.
%     Subsequently, 
%     we generate multiple linear targets with varying complexity measures.
%     These targets are then trained using CNNs with varying numbers of channels $M$,
%     and we plot the training error against $M$.
%     } \label{fig1}
% \end{figure}

\section{Conclusion}
In practical machine learning applications, approximation results provide useful insights into model selection.
Temporal CNNs are adapted to approximate sequential relationships which have regular spectrum (small $C_1$) and decaying memory (small $C_2$), 
that is, 
we have a rate estimate as in \cref{thm:jackson} if and only if the target has small complexity measure $C_1$ and $C_2$. 
This highlights the importance of both Jackson-type and Bernstein-type results in machine learning, 
as they enable us to assess the suitability of an architecture prior to investing computational resources and time in a trial-and-error process.
By doing so, we can increase the efficiency and effectiveness of the model selection process.
% \input{experiments.tex}

% \newpage
% \input{backup.tex}

\bibliographystyle{packages/splncs04}
\bibliography{bibliography}
\renewcommand*{\proofname}{Proof}
\newpage
\appendix
\section{Appendix}
\subsection{Tensor structure of the dilated convolution architectures.}
In this section we discuss the tensor structure related with the dilated convolution architectures $\Hcnnl$. 
Recall the model representation have the following form
\begin{equation}
    \rhh_i
    =
    \sum_{i_1, \dots, i_{K}=1}^M
    \bm{w}_{K-1, i_{K-1}, i_K}
    \Conv{}_{l^{K-1}}
    % \bm{w}_{K-2, i_{K-2}, i_{K-1}}
    % \Conv{}_{l^{K-2}}
    \dots
    \Conv{}_{l^1}
    \bm{w}_{0, i, i_1},
\end{equation}

We define the following tensorization of a vector $T:\R^{l^K} \to \R^{l\times\dots\times l}$ such that
\begin{align}
        [T(\bm \rho)]_{a_1,\dots,a_K}
        =
        \rho
        \left(
            \sum_{j=1}^{K}
            a_j l^{j-1}
        \right),
        \qquad
        a_j \in \{0, \dots, l-1\}.
\end{align}
This means that the value at position $t$ is mapped to position $(a_1,\dots,a_K)$ in the tensor, where $a_1\dots a_K$ is the base $l$ expansion of $t$.
\begin{proposition}
    For the representation $\rhh_i$  of the model, we have 
    \begin{align}
        T(\rhh_i) = \sum_{i_1, \dots, i_{K}=1}^M
        \bm{w}_{K-1, i_{K-1}, i_K}
        \otimes
        \bm{w}_{K-2, i_{K-2}, i_{K-1}}
        \otimes
        \dots
        \otimes
        \bm{w}_{0, i, i_1}.
    \end{align}
    \begin{proof}
        It is sufficient to consider  $\rho_{K} = \bm w_{K-1} \Conv{}_{l^{K-1}} \cdots \Conv{}_{l^1}\bm w_{0}$ and $T(\bm\rho)$.
        Let $\bm A = \bm w_{K-1} \otimes \cdots \otimes\bm w_{0}$, we have 
        \begin{align}
            [\bm A]_{a_{K-1}\dots a_0}=  w_{K-1}(a_{K-1})\cdot w_{K-2}(a_{K-2} )\cdot \, \cdots \, \cdot w_{0}(a_{0})
        \end{align}
        We now prove $[\bm A]_{a_{K-1}\dots a_0} = \rho_K
        \left(
            \sum_{j=0}^{K-1}
            a_{j} l^{j}
        \right)$ by induction.
        When $K=2$ we have $\bm A$ being a $l\times l$ matrix, and 
        \begin{align}
            \rho_2(t) &= (\bm w_{1}\Conv{}_l \bm w_0 )(t) =  \sum_{s\geq 0}w_1(s) \cdot w_0(t-l s).
        \end{align}
        Substitute in we get
        \begin{align}
            \rho_2\left(
                a_0 + l a_1
            \right) & = \sum_{s=0}^{l-1}w_1(s) \cdot w_0(a_0 + l (a_1- s)).
        \end{align}
        Note that the  domain of $\bm w_0$ is $[0, l-1]$, this means that $a_0 + l (a_1- s)$ is in the domain if and only if $s = a_1$, where we have
        \begin{align}
            \rho_2\left(
                a_0 + l a_1
            \right) & = w_1(a_1) \cdot w_0(a_0).
        \end{align}
        This proves the case when $K=2$.
        Now, suppose the it holds for $K=k$. Then, for $K=k+1$ we have
        \begin{align}
            \bm \rho_{k+1} = \bm w_{k} \Conv{}_{l^k}\bm\rho_{k}.
        \end{align}
        Thus,
        \begin{align}
            \rho_{k+1}(t)&= \sum_{s=0}^{l-1}w_{k}(s)\rho_{k}(t-l^k s) \\
            & = \sum_{s=0}^{l-1}w_{k}(s)\rho_{k}
            \left( \sum_{j=0}^{k-1} a_{j} l^{j}+l^k (a_k - s) \right).
        \end{align}
        Domain of $\rho_{k}$ is $[0,l^k]$, which means $s=a_K$
        \begin{align}
            \rho_{k+1}\left( \sum_{j=0}^{k} a_{j} l^{j}\right) = w_k(a_k)\rho_k\left( \sum_{j=0}^{k-1} a_{j} l^{j}\right).
        \end{align}
            By the induction hypothesis we have
            \begin{align}
                \rho_{k+1}\left( \sum_{j=0}^{k} a_{j} l^{j}\right)& = w_k(a_k)\, w_{k-1}(a_{k-1})\cdot w_{k-2}(a_{k-2} )\cdot \, \cdots \, \cdot w_{0}(a_{0})\\
                & = [\bm A]_{a_k\dots a_0}.
            \end{align}
            This holds for all $k>2$ which ends the proof.
    \end{proof}
\end{proposition}
This proposition show that the dilated convolution exhibits a special tensor structure.
We can tensorize the target in the same manner, which allowed us to examine how it can be approximated.

We now present the proof of the main theorems.

\begin{proof}[\textbf{Jackson-type results}]
    Suppose we use a model with $K$ layers and $M$ channels.
    For $\bm x \in \X$  with $\norm{\bm x}\leq 1$, we have
    \begin{align}\label{eq:two parts error}
        \abs{H_t(\bm x)  - \hat H_t(\bm x)}^2 & = \abs{\sum_{s=0}^\infty {\rh}(s)^\top x(t-s) -  {\rhh}(s)^\top x(t-s)}^2\\
        &\leq  \sum_{s=0}^\infty\abs{ {\rh}(s)^\top -  {\rhh}(s)^\top}^2\sum_{s=0}^\infty\abs{x(t-s)}^2  \\
        & \leq  \sum_{s=0}^\infty\abs{ {\rh}(s)^\top -  {\rhh}(s)^\top}^2 \\
        \intertext{Note that when $s\geq l^K$ , $\rhh(s)=0$.}
        &= \label{eq:error equality}
         \sum_{i=1}^d \sum_{s=0}^{l^K-1}\Big\rvert \rh_i(s) - \rhh_i(s)\Big\rvert^2
        + \sum_{s=l^K}^{\infty} \Big\rvert \rh(s) \Big\rvert ^2 \\
        & =  \sum_{i=1}^d \norm{T(\rh_i) - T(\rhh_i)}_F^2 + \sum_{s=l^K}^{\infty} \Big\rvert \rh(s) \Big\rvert ^2
    \end{align}
    For the first term we have
    \begin{align}
        T(\rh_i) - T(\rhh_i) & = \sum_{i_1,i_2,...,i_K=1}^l s_{i_1\dots i_K} \bm u_{i_1} \otimes \cdots \otimes\bm u_{i_K} - 
        \sum_{i=1}^{M} [\hat s_{i} \hat {\bm u} _{1i}]\otimes \cdots \otimes \hat{ \bm u}_{Ki} \\
        & =  \sum_{i=1}^{l^K}s_i \bm u_{1i}\otimes \cdots \otimes u_{iK} - \sum_{i=1}^{M} [\hat s_{i} \hat {\bm u} _{1i}]\otimes \cdots \otimes \hat{ \bm u}_{Ki}.
    \end{align}
    By setting $\hat{\bm u}_{ki} = \bm u_{ki}$ and $\hat s_i = s_i$ for $i=1\dots M$, we have
    \begin{align}
        T(\rh_i) - T(\rhh_i) & =\sum_{i=M+1}^{l^K}s_i \bm u_{1i}\otimes \cdots \otimes u_{iK}.
    \end{align}

    By the orthogonality of the HOSVD bases we have
    \begin{align}
        \norm{ T(\rh_i) - T(\rhh_i) }_F^2 = \sum_{i=M+1}^{l^K}\abs{s_i} ^2.
    \end{align}
    Thus, by the definition of the complexity measure we have
    \begin{align}
        \sum_{i=1}^d \norm{T(\rh_i) - T(\rhh_i)}_F^2 \leq C_1\depen{g}(\bm H)g(M).
    \end{align}

    For the second term we have 
    \begin{align}
        \sum_{s=l^K}^{\infty} \Big\rvert \rh(s) \Big\rvert ^2 \leq C_{2}\depen{f}(\bm H) f(l^K) 
    \end{align}
    directly from the definition of $C_2$.
    This completes the proof.
\end{proof}

We next present the proof of the inverse approximation.

\begin{proof}[\textbf{Bernstein-type results}]
By the assumption for all $t$ we have
\begin{align}
    \mathop{\E}_{\bm x \sim \mathcal D} [|H_t(\bm x) - \hat H_t(\bm x)|^2]\leq Ag(M) + Bf(l^K)
\end{align}
holds for all $M,K$.
Where $\mathcal D$ is a distribution such that at every time step $t$ we have $x(t)\sim \mathcal N(0,1)$.
We then have 
\begin{align}
    \mathop{\E}_{\bm x \sim \mathcal D} [|H_t(\bm x) - \hat H_t(\bm x)|^2] & =  \mathop{\E}_{\bm x \sim \mathcal D}\left [\abs{\sum_{s=0}^\infty {\rh}(s)^\top x(t-s) -  {\rhh}(s)^\top x(t-s)}^2\right ]\\
    & = \sum_{s=0}^\infty\abs{\rh(s)-\rhh(s)}^2.
\end{align}
Now  the assumption becomes for all $M,K$
\begin{align}
    \sum_{s=0}^\infty\abs{\rh(s)-\rhh(s)}^2 \leq Ag(M) + Bf(l^K).
\end{align}
Continue from \cref{eq:error equality} we have
\begin{align}
   \sum_{s=0}^\infty\abs{ {\rh}(s)^\top -  {\rhh}(s)^\top}^2 
        &=
         \sum_{i=1}^d \sum_{s=0}^{l^K-1}\Big\rvert \rh_i(s) - \rhh_i(s)\Big\rvert^2
        + \sum_{s=l^K}^{\infty} \Big\rvert \rh(s) \Big\rvert ^2.
\end{align}
Thus, for all $K, M$ we have
\begin{align}
    \sum_{i=1}^d \sum_{s=0}^{l^K-1}\Big\rvert \rh_i(s) - \rhh_i(s)\Big\rvert^2
    + \sum_{s=l^K}^{\infty} \Big\rvert \rh(s) \Big\rvert ^2 \leq Ag(M) + Bf(l^K).
\end{align}
Recall that $\rhh \in \ell^2$ and $g,f$ converge to zero.
Take $M$ goes to infinity we have
\begin{align}
    \sum_{s=l^K}^{\infty} \Big\rvert \rh(s) \Big\rvert ^2 \leq Bf(l^K).
\end{align}
Which implies that $C_2\depen{f}(\bm H) \leq B$.
Take $K$ goes to infinity, and take the model to match the bases of $T(\rh)$ we have
\begin{align}
  \sum_{j=1}^d\sum_{s=M+1}^\infty |s_{i}\depen{j,K}|^2 =  \sum_{i=1}^d \sum_{s=0}^{l^K-1}\Big\rvert \rh_i(s) - \rhh_i(s)\Big\rvert^2
   \leq Ag(M),
\end{align}
which implies $C_1\depen{\bm H} \leq A$.
\end{proof}

\end{document}